%% file: main.tex
\documentclass[11pt]{article}
\usepackage{coling2018}
\usepackage{times}
\usepackage{url}
\usepackage{latexsym}

\def\figref#1{Fig.~\ref{#1}}
\def\secref#1{Sec.~\ref{#1}}

\usepackage{graphicx}
\usepackage{color}
\usepackage{amsmath}
\usepackage{tabularx}
\usepackage{wrapfig}
\usepackage{arydshln}
\graphicspath{ {figs/} }

\title{Dynamic Multi-Level Multi-Task Learning for Sentence Simplification}

\author{Han Guo \;\;\;\;\;\;\; Ramakanth Pasunuru \;\;\;\;\;\;\; Mohit Bansal \\
  UNC Chapel Hill \\
  {\tt \{hanguo, ram, mbansal\}@cs.unc.edu} \\
 }

\date{}

\begin{document}
\maketitle

\begin{abstract}
Sentence simplification aims to improve readability and understandability, based on several operations such as splitting, deletion, and paraphrasing. However, a valid simplified sentence should also be logically entailed by its input sentence. In this work, we first present a strong pointer-copy mechanism based sequence-to-sequence sentence simplification model, and then improve its entailment and paraphrasing capabilities via multi-task learning with related auxiliary tasks of entailment and paraphrase generation. Moreover, we propose a novel `multi-level' layered soft sharing approach where each auxiliary task shares different (higher versus lower) level layers of the sentence simplification model, depending on the task's semantic versus lexico-syntactic nature. We also introduce a novel multi-armed bandit based training approach that dynamically learns how to effectively switch across tasks during multi-task learning. Experiments on multiple popular datasets demonstrate that our model outperforms competitive simplification systems in SARI and FKGL automatic metrics, and human evaluation. Further, we present several ablation analyses on alternative layer sharing methods, soft versus hard sharing, dynamic multi-armed bandit sampling approaches, and our model's learned entailment and paraphrasing skills. 
\end{abstract}

\input{introduction.tex}

\input{related_works.tex}

\input{models.tex}

\input{setup.tex}

\input{results.tex}

\input{analysis.tex}

\input{conclusions.tex}

\section*{Acknowledgments}
We thank the reviewers for their helpful comments (and Xingxing Zhang for providing preprocessed datasets). This work was supported by DARPA (YFA17-D17AP00022), Google Faculty Research Award, Bloomberg Data Science Research Grant, and Nvidia GPU awards. The views contained in this article are those of the authors and not of the funding agency.

\bibliographystyle{acl}
\bibliography{citations}

\appendix
\input{appendix}

\end{document}

%% file: introduction.tex
\section{Introduction}
\label{sec-intro}

\blfootnote{
\hspace{-0.65cm}  
This work is licensed under a Creative Commons Attribution 4.0 International License. License details: \url{http://creativecommons.org/licenses/by/4.0/}.
}

Sentence simplification is the task of improving the readability and understandability of an input text. This challenging task has been the subject of research interest because it can address automatic ways of improving reading aids for people with limited language skills, or language impairments such as dyslexia~\cite{rello2013impact}, autism~\cite{evans2014evaluation}, and aphasia~\cite{carroll1999simplifying}. It also has wide applications in NLP tasks as a preprocessing step, for example, to improve the performance of parsers~\cite{chandrasekar1996motivations}, summarizers~\cite{klebanov2004text}, and semantic role labelers~\cite{vickrey2008sentence,woodsend2014text}.

Several sentence simplification systems focus on operations such as splitting a long sentence into shorter sentences~\cite{SIDDHARTHAN2006SyntacticSA,petersen2007text}, deletion of less important words/phrases~\cite{knight2002summarization,clarke2006models,filippova2008dependency}, and paraphrasing~\cite{devlin1999simplifying,inui2003text,kaji2002verb}. Inspired from machine translation based neural models, recent work has built end-to-end sentence simplification models along with attention mechanism, and further improved it with reinforcement-based policy gradient approaches~\cite{zhang2017dress}. Our baseline is a novel application of the pointer-copy mechanism~\cite{see2017get} for the sentence simplification task, which allows the model to directly copy words and phrases from the input to the output. We  further improve this strong baseline by bringing in auxiliary entailment and paraphrasing knowledge via soft and dynamic multi-level, multi-task learning.\footnote{All code and pretrained models available at:~\url{https://github.com/HanGuo97/MultitaskSimplification}.} 

Apart from the three simplification operations discussed above, we also ensure that the simplified output is a directed logical entailment w.r.t. the input text, i.e., does not generate any contradictory or unrelated information. We incorporate this entailment skill via multi-task learning~\cite{luong2015multi} with an auxiliary entailment generation task. 
Further, we also induce word/phrase-level paraphrasing knowledge via a paraphrase generation task, enabling parallel learning of these three tasks in a three-way multi-task learning setup. 
We employ a novel `multi-level' layered, soft sharing approach, where the parameters between the tasks are loosely coupled at different levels of layers; we share higher-level semantic layers between the sentence simplification and entailment generation tasks (which teaches the model to generate outputs that are entailed by the full input), while sharing the lower-level lexico-syntactic layers between the sentence simplification and paraphrase generation tasks (which teaches the model to paraphrase only the smaller sub-sentence pieces). 

Finally, we also propose a multi-armed bandit approach that dynamically learns an effective schedule (curriculum) of switching between tasks for optimization during multi-task learning, instead of the traditional approach with a manually-tuned, static (fixed) mixing ratio~\cite{luong2015multi}. This dynamic approach allows us to achieve not only  equal, but in fact better results than the manual approach, while importantly avoiding the hassle of tuning on the large space of mixing ratios over several different tasks.

Empirically, we evaluate our system on three standard datasets: Newsela, WikiSmall, and WikiLarge. First, we show that our pointer-copy baseline is significantly better than sequence-to-sequence models, and competitive w.r.t. the state-of-the-art. Next, we show that our multi-level, multi-task framework performs significantly better than our strong pointer baseline and other competitive sentence simplification models on both automatic evaluation as well as on human study simplicity criterion. Further, we show that the dynamic multi-armed bandit based switching of tasks during training improves over the traditional manually-tuned static mixing ratio. Lastly, we show several ablation studies based on different layer-sharing approaches (higher versus lower) with auxiliary tasks, hard versus soft sharing, dynamic mixing ratio sampling, as well as our model's learned entailment and paraphrasing skills.

%% file: related_works.tex
\section{Related Work}
\label{section:related-works}

Previous approaches to sentence simplification systems range from hand-designed rules~\cite{SIDDHARTHAN2006SyntacticSA}, to syntactic and lexical simplification via synonyms and paraphrases~\cite{Siddharthan2014ASO,kaji2002verb,horn2014learning,glavavs2015simplifying}, as well as treating simplification as a monolingual MT task, where operations are learned from examples of complex-simple sentence pairs~\cite{specia2010translating,koehn2007moses,coster2011learning,Zhu2010AMT,Wubben2012SentenceSB,Narayan2014HybridSU}. 
Recently,~\newcite{Xu2016OptimizingSM} trained a syntax-based MT model using the newly proposed SARI as a simplification-specific objective. Further,~\newcite{zhang2017dress} used reinforcement learning in a sequence-to-sequence approach to directly optimize simplification metrics. In this work, we first introduce the pointer-copy mechanism~\cite{see2017get} as a novel application to sentence simplification, and then use multi-task learning to bring in auxiliary entailment and paraphrasing skills.

Multi-task learning, known for improving the generalization performance of a task with related tasks, has successful application to many domains of machine learning~\cite{caruana1998multitask,collobert2008unified,girshick2015fast,luong2015multi,pasunuru2017multitask,Pasunuru2017TowardsIA}. Although there are many variants of multi-task learning~\cite{ruder2017sluice,hashimoto2017ajm,luong2015multi}, 
our approach is similar to~\newcite{luong2015multi}, where different tasks share some common model parameters with alternating mini-batches optimization. 
In this work, we explore a multi-level (i.e., task-specific higher-level semantic versus lower-level lexico-syntactic layer sharing) and soft-sharing mechanism for improving sentence simplification via related tasks of entailment and paraphrase generation.

Recognizing Textual Entailment (RTE) is the task of predicting entailment, contradiction, or neutral relationships, and is useful for many downstream tasks like Q\&A, summarization, and information retrieval~\cite{harabagiu2006methods,dagan2006pascal,lai2014illinois,jimenez2014unal}. Neural network models~\cite{bowman2016snli,parikh2016decomposable} and large datasets~\cite{bowman2016snli,williams2017broad} enabled recent strong progress. Recently,~\newcite{Pasunuru2017TowardsIA} and~\newcite{han2017multitask} presented results using entailment generation as an auxiliary task for abstractive summarization; however, we use entailment as well as paraphrasing knowledge in a soft and multi-level layer sharing setup to improve sentence simplification.

Previous work~\cite{barzilay2001extracting,ganitkevitch2013ppdb,Wieting2017PushingTL} has developed methods and datasets for generating paraphrase pairs which can be useful for downstream applications such as question answering, semantic parsing, and information extraction~\cite{fader2013paraphrase,berant2014semantic,zhang2015exploiting}. \newcite{Wieting2017PushingTL} recently introduced a large sentential paraphrase dataset via back-translation, and showed promising results when applied to learning sentence embeddings. In this work, we use this paraphrase dataset as an auxiliary generation task to improve our sentence simplification model by teaching it about paraphrasing in a multi-task setting.

Many control problems can be cast as a multi-armed bandits algorithm, where the goal of the agent is to select the arm/action from one of the $M$ choices that gives the maximum expected future reward~\cite{bubeck2012regret}. Optimal control and reinforcement learning have been used to find the trade-off between exploitation and exploration, and yield theoretically-sound regret bounds, e.g., Boltzmann exploration~\cite{kaelbling1996reinforcement}, UCB~\cite{auer2002finite}, Thompson sampling~\cite{chapelle2011empirical}, adversarial bandits~\cite{auer2002nonstochastic}, and information gain using variational approaches~\cite{houthooft2016vime}. Recently, \newcite{graves2017automated} use a non-stationary multi-armed bandit to automatically select the curriculum or syllabus that a neural network follows so as to maximize learning efficiency. \newcite{sharma2017online} use multi-armed bandit sampling to choose which domain data (harder vs. easier) to feed as input to a single model (using different Atari games), whereas we use multi-armed bandit sampling to decide the optimization curriculum (mixing ratio) among our three models for sentence simplification, entailment generation, and paraphrase generation (with different softly-shared layers).

%% file: models.tex
\section{Models}
\label{sec:models}

In this section, we first describe our sentence simplification baseline model with attention mechanism, which is further improved by pointer-copy mechanism. Later, we introduce our two auxiliary tasks (entailment and paraphrase generation) and discuss how they can share specific lower/higher-level layers/parameters to improve the sentence simplification task in a multi-task learning setting. Finally, we discuss our new multi-armed bandit based dynamic multi-task learning approach.

\subsection{Pointer-Copy Baseline Sentence Simplification Model}
Our baseline is a 2-layer sequence-to-sequence model with both attention~\cite{bahdanau2014attention} and pointer-copy mechanism~\cite{see2017get}. Given the sequence of input/source tokens $x = \{x_1, ..., x_{T_s}\}$, the model learns an auto-regressive distribution over output/target tokens $y = \{y_1, ..., y_{T_o}\}$, which is defined as $P_{vocab}(y \vert x; \theta) = \prod_t p(y_t \vert y_{1:t-1}, x; \theta)$, 
where $\theta$ represents model parameters and $p(y_t \vert y_{1:t-1}, x; \theta)$ is probability of generating token $y_t$ at decoder time step $t$ given the previous generated tokens $y_{1:t-1}$ and input $x$. Given encoder hidden states $\{h_i\}$, and decoder's $t^{\text{th}}$ time step hidden state (of last layer) $s_t$, the context vector $c_t=\sum_i \alpha_{t,i} h_i$, where the attention weights $\alpha_{t,i}$ define an attention distribution over encoder hidden states: $\alpha_{t,i} = \exp(e_{t,i})/\sum_k\exp(e_{t,k})$, where $e_{t,i} = v_a^T \tanh(W_a s_t + U_a h_i + b_a)$.
Finally, the conditional distribution at each time step $t$ of the decoder is defined as $p(y_t|y_{1:t-1},x;\theta) = \textrm{softmax}(W_s s'_t)$, where the final hidden state $s'_t$ is a combination of context vector $c_t$ and last layer hidden state $s_t$ and is defined as $s'_t = \tanh(W_c[c_t,s_t])$, where $W_s$ and $W_c$ are trained parameters.

\noindent\textbf{Pointer-Copy Mechanism:} This helps in directly copying the words from the source inputs to the target outputs via merging the generative distribution and attention distribution (as a proxy of copy distribution). The goal of sentence simplification is to rewrite sentences more simply, while preserving important information; hence, it also involves significant amount of copying from the source. Our pointer mechanism approach is similar to~\newcite{see2017get}. At each time step of the decoder, the model makes a (soft) choice between words from the vocabulary distribution $P_{vocab}$ and attention distribution $P_{att}$ (based on words in the input) using the word generation probability $p_g = \sigma (W_g c_t + U_g s_t + V_g d_t + b_g)$, where $\sigma (\cdot)$ is sigmoid, $W_g, U_g, V_g$ and $b_g$ are trainable parameters, and $d_t$ is decoder input. The final vocabulary distribution is defined as the weighted combination of vocabulary and attention distributions:
\vspace{-5pt}
\begin{equation}
P_f (y) = p_g P_{vocab} (y) + (1 - p_g) P_{att} (y)
\vspace{-5pt}
\end{equation}

\subsection{Auxiliary Tasks}

\paragraph{Entailment Generation} 
The task of entailment generation is to generate a hypothesis which is entailed by the given input premise. A good simplified sentence should be entailed by (follow from) the source sentence, and hence we incorporate such knowledge through an entailment generation task into our sentence simplification task. We share the higher-level semantic layers between the two tasks (see reasoning in Sec.~\ref{subsec:multi-level-sharing-mechanism} below). We use entailment pairs from SNLI~\cite{bowman2016snli} and Multi-NLI~\cite{williams2017broad} datasets for training our entailment generation model, where we use the same architecture as our sentence simplification model.

\paragraph{Paraphrase Generation} 
Paraphrase generation is the task of generating similar meaning phrases or sentences by reordering and modifying the syntax and/or lexicon. Paraphrasing is one of the common operations used in sentence simplification, i.e, by substituting complex words and phrases with their simpler paraphrase forms. Hence, we add this knowledge to the sentence simplification task via multi-task learning, by sharing the lower-level lexico-syntactic layers between the two tasks (see reasoning in Sec.~\ref{subsec:multi-level-sharing-mechanism} below). For this, we use the paraphrase pairs from ParaNMT~\cite{Wieting2017PushingTL}. Here, again, we use the same architecture as our sentence simplification model.

\begin{figure}
\centering
\begin{minipage}[t]{.48\linewidth}
\centering
\includegraphics[width=0.98\linewidth]{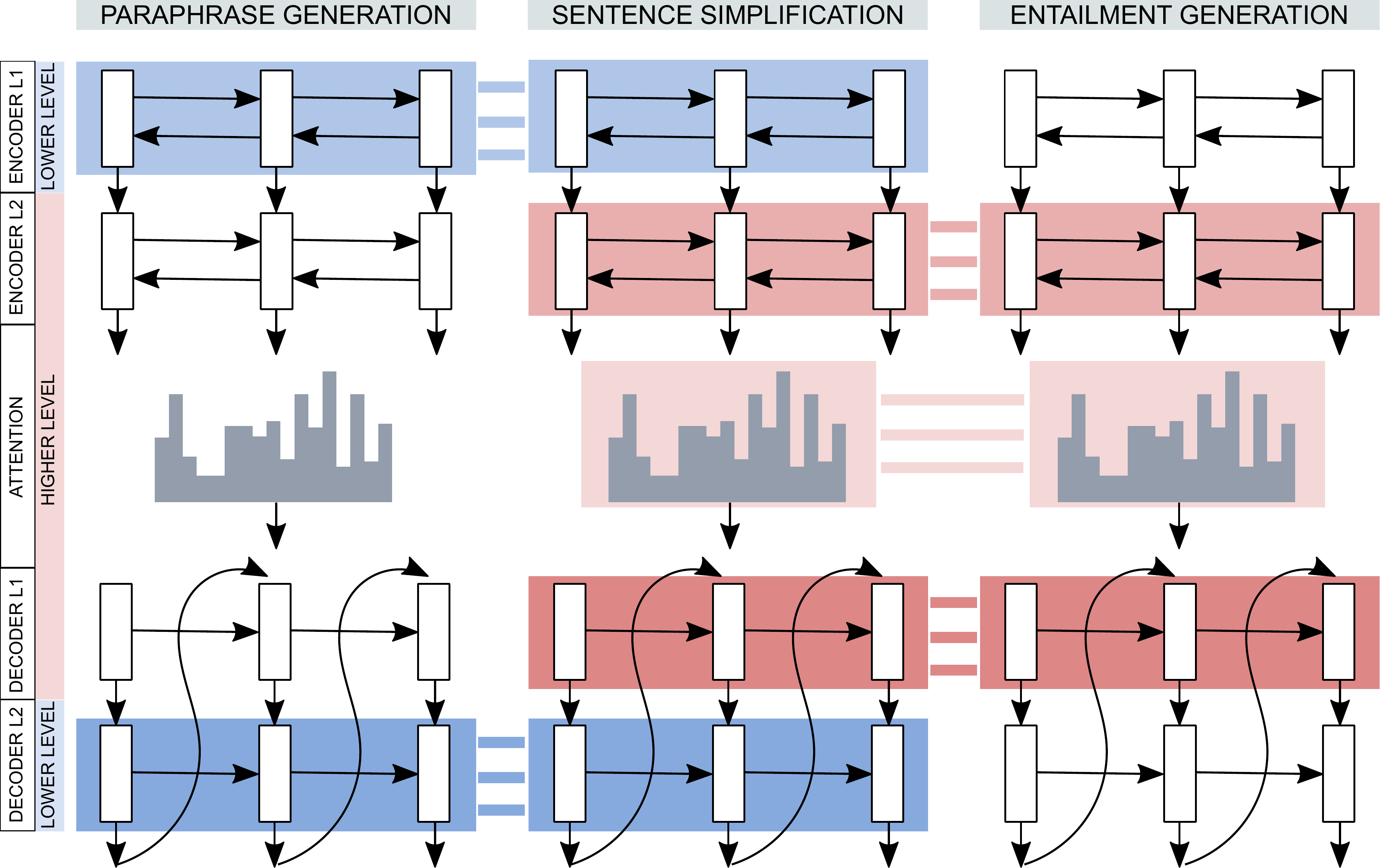}
\vspace{-10pt}
\caption{Overview of our 3-way multi-task model. Same color and dashed connections represent soft-shared parameters in different layers.
}
\vspace{-10pt}
\label{fig:multitask}
\end{minipage}
\hfill
\begin{minipage}[t]{.48\linewidth}
\centering
\includegraphics[width=0.98\linewidth]{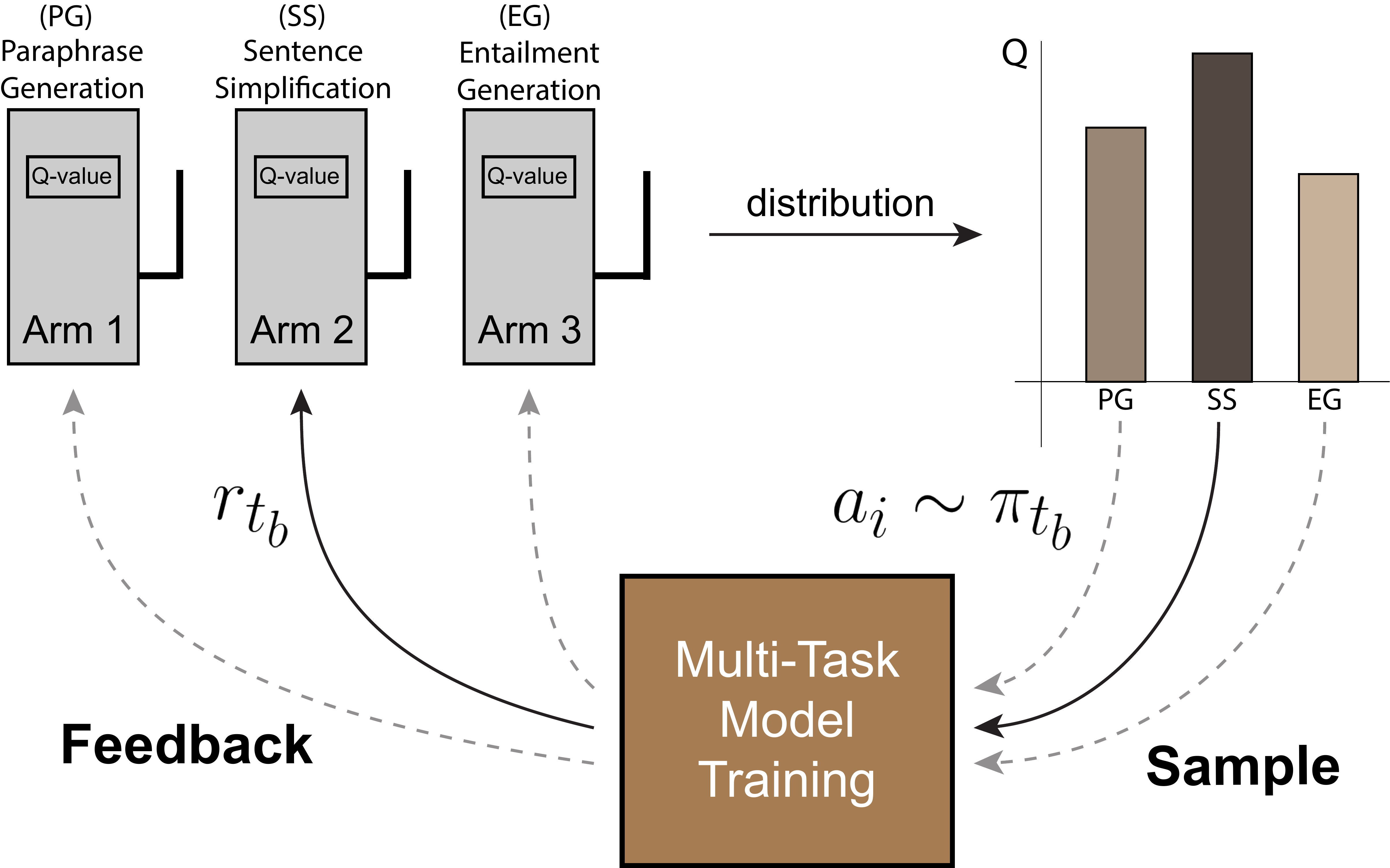}
\vspace{-10pt}
\caption{Overview of our multi-armed bandits algorithm for dynamic mixing ratio learning. It consists of a controller with 3 arms/tasks.
}
\vspace{-10pt}
\label{fig:mab}
\end{minipage}
\end{figure}

\subsection{Multi-Task Learning}
\label{subsec:multi-level-sharing-mechanism}
In this subsection, we discuss our multi-task, multi-level soft sharing strategy with parallel training of sentence simplification and related auxiliary tasks (entailment and paraphrase generation). 

The predominant approach for multi-task learning in sequence-to-sequence models is to directly hard-share all encoder/decoder layers/parameters~\cite{luong2015multi,johnson2016google,pasunuru2017multitask,kaiser2017one}. However, this approach places very strong constraints/priors on the primary model to compress knowledge from diverse tasks. We believe that while the auxiliary tasks considered in this work share many similarities with the primary sentence simplification task, they are still different in either lower-level or higher-level representations (e.g., entailment will deal with higher-level, full-sentence logical inference, while paraphrasing will handle the lower-level intermediate word/phrase simplifications). 
In this section, we propose to relax the priors in two ways: (1) we share the model parameters in a finer-grained scale, i.e. layer-specific sharing, by keeping some of their parameters private, while sharing related representations; and (2) we encourage shared parameters to be close in certain distance metrics with a penalty term instead of hard-parameter-tying~\cite{luong2015multi}.

\paragraph{Multi-Level Sharing Mechanism} Fig.~\ref{fig:multitask} shows our multi-task model with parallel training of three tasks: sentence simplification (primary task), entailment generation (auxiliary task), and paraphrase generation (auxiliary task). 
Recently,~\newcite{belinkov2017neural} observed that different layers in a sequence-to-sequence model (trained on translation) exhibit different functionalities: lower-layers (closer to inputs) of the encoder learn to represent word structure while higher layers (farther from inputs) are more focused on semantics and meanings (\newcite{zeiler2014visualizing} observed similar findings for convolutional image features). Based on these findings, we share the higher-level layers\footnote{We found that sharing higher-level semantic layers (farther from input/output), i.e., encoder layer 2, attention, and decoder layer 1 (in~\figref{fig:multitask}), to work well. See Sec.~\ref{sec:analysis} for ablations on alternative layer sharing methods.} between the entailment generation and sentence simplification tasks, since they share higher semantic-level language inference skills (for full sentence-to-sentence logical directedness). On the other hand, we share the lower-level lexico-syntactic layers\footnote{We found that sharing lower-level lexico-syntactic layers (closer to input/output), i.e., encoder layer 1 and decoder layer 2 (in~\figref{fig:multitask}), to work well. See Sec.~\ref{sec:analysis} for ablations on alternative layer sharing methods.} between the paraphrase generation and sentence simplification tasks, since they share more word/phrase and syntactic level paraphrasing knowledge to simplify the smaller, intermediate sentence pieces.
Sec.~\ref{sec:analysis} present empirical ablations to support our intuitive layer sharing.\footnote{Note that even though entailment just tries to generate shorter, logical-subset sub-sentences, the overall saliency and quality of the simplified output is still balanced because the entailment task is flexibly (softly) shared with the paraphrasing and sentence simplification tasks, and the final model mixture is chosen based on simplification task metrics (see output examples in Fig.~\ref{fig:examples} where our multi-task model generates entailed sentences with important information).}

\paragraph{Soft Sharing}
In multi-task learning, we can do either hard sharing or soft sharing of parameters. Hard sharing directly ties the parameters to be shared, and receives gradient information from multiple tasks. On the other hand, soft sharing only loosely couples the parameters, and encourages them to be close in representation space. Hence the soft sharing approach gives more flexibility for parameter sharing, hence allowing different tasks to choose what parts of their parameters space to share. We minimize the $l_2$ distance between shared parameters as a regularization along with the cross entropy loss. Hence, the final loss function of the primary task with a related auxiliary task is defined as follows:
\vspace{-6pt}
\begin{equation}
L(\theta) = - \log P_f (y|x;\theta) + \lambda || \theta_s - \phi_s ||
\vspace{-6pt}
\end{equation}
where $\theta$ represents the full parameters of the primary task (sentence simplification), $\theta_s$ and $\phi_s$ are the subsets of shared parameters between the primary and auxiliary task resp., and $\lambda$ is a hyperparameter.

\paragraph{Multi-Task Training}
We employ multi-task learning with parallel training of related tasks in alternate mini-batches based on a mixing ratio $\alpha_{ss}{:}\alpha_{eg}{:}\alpha_{pp}$, where we alternatively optimize $\alpha_{ss}$, $\alpha_{eg}$, $\alpha_{pp}$ mini-batches of sentence simplification, entailment generation, and paraphrase generation, respectively, until all models converge. In the next section, we discuss a new approach to replace this static mixing ratio with dynamically-learned task switching.

\subsection{Dynamic Mixing Ratio Learning}
\label{subsec:active-sampling}
Current multi-task models are trained via alternate mini-batch optimization based on a task `mixing ratio'~\cite{luong2015multi,pasunuru2017multitask}, i.e., how many iterations on each task relative to other tasks (see end of Sec.~\ref{subsec:multi-level-sharing-mechanism}). This is usually treated as a very important hyperparameter to be tuned, and the search space scales exponentially with the number of tasks. Hence, we importantly replace this manually-tuned and static mixing ratio with a `dynamic' mixing ratio learning approach, where a controller automatically switches between the tasks during training, based on the current state of the multi-task model. Specifically, we use a multi-armed bandits based controller with Boltzmann exploration~\cite{kaelbling1996reinforcement} with an exponential moving average update rule.

We view the problem of learning the right mixing of tasks as a sequential control problem, where the controller's goal is to decide the next task/action after every $n^{s}$ training steps in each task-sampling round $t_b$.\footnote{We set $n^{s}$ to $10$ to reduce variance of estimates, i.e., the bandit controller's task/action will be trained for $10$ mini-batches.} Let $\{a_1, ..., a_M\}$ represent the set of 3 tasks in our multi-task setting, i.e., sentence simplification, entailment generation, and paraphrase generation. We model the controller as a $M$-armed bandits, where it selects a sequence of actions/arms over the current training trajectory to maximize the expected future payoffs (see Fig.~\ref{fig:mab}). At each round $t_b$, the controller selects an arm based on noisy value estimates and observes rewards $r_{t_b}$ for the selected arm (we use the  negative validation loss of the primary task as the reward in our setup). One problem in bandits learning is the trade-off between exploration and exploitation, where the agent needs to make a decision between taking the action that yields the best payoff on current estimates, or explore new actions whose payoffs are not yet certain. For this, we use the Boltzmann exploration~\cite{kaelbling1996reinforcement} with exponentially moving action value estimates. Let $\pi_{t_b}$ be the policy of the bandit controller at round $t_b$, we define this to be:
\vspace{-9pt}
\begin{equation}
\pi_{t_b}(a_{i})  = \exp(Q_{t_b,i} / \tau) \Big/ \sum_{j=1}^M\exp(Q_{t_b,j} / \tau)
\vspace{-7pt}
\end{equation}
where $Q_{t_b,i}$ is the estimated action value of each arm $i$ at round $t_b$, and $\tau$ is the temperature.\footnote{We tried decaying the temperature variable, but we didn't find this to very beneficial, so we instead fix this to $1.0$.} If $Q_{0,i}$ is the initial value estimate of arm $i$, then $Q_{t_b,i}$ is the exponentially weighted mean with the decay rate $\alpha$:
\vspace{-9pt}
\begin{equation}
Q_{t_b,i} = (1-\alpha)^{t_b} Q_{0,i} + \sum_{k=1}^{t_b}\alpha(1-\alpha)^{t_b-k}r_k
\vspace{-9pt}
\end{equation}
To further help the exploration process, we follow the principle of optimism under uncertainty~\cite{sutton1998reinforcement} and set $Q_{0,i}$ to be above the maximum empirical rewards. Empirically, we show that this approach of `dynamic mixing ratio' is equal or better than the traditional static mixing ratio (see Table~\ref{table:dynamic-mixing-results}).
Also, we further show ablation study in Sec.~\ref{sec:analysis} to show that this switching approach is better than the alternative approach of first using multi-armed bandits for finding an optimal `final' mixing ratio and then re-training the model based on this bandits-selected mixing ratio.

%% file: setup.tex
\section{Evaluation Setup}
\label{section:setup}

\paragraph{Datasets}
\label{subsec:datasets}
We first describe the three standard sentence simplification datasets we evaluate on: Newsela, WikiSmall, and WikiLarge; next, we describe datasets for our auxiliary entailment and paraphrase generation tasks. \textbf{Newsela}~\cite{xu2015problems} is acknowledged as a higher-quality dataset for studying sentence simplifications, as opposed to Wikipedia-based datasets which automatically align complex-simple sentence pairs and have generalization issues~\cite{zhang2017dress,xu2015problems,amancio2014analysis,hwang2015aligning,vstajner2015deeper}. Newsela consists of $1,130$ news articles, 
and we follow previous work~\cite{zhang2017dress} to use the first $1,070$ documents for training, and $30$ documents each for development and test.
\textbf{WikiSmall}~\cite{Zhu2010AMT} contains automatically-aligned complex-simple sentences from the ordinary-simple English Wikipedias. The data has $89,042$ pairs for training and $100$ for test. We use the $205$-pairs validation set from~\newcite{zhang2017dress}.
\textbf{WikiLarge}~\cite{zhang2017dress} is a larger Wikipedia corpus aggregating pairs from~\newcite{Kauchak2013ImprovingTS},~\newcite{Woodsend2011LearningTS}, and WikiSmall. We use the exact training/evaluation sets provided by~\newcite{zhang2017dress}.
\textbf{SNLI and MultiNLI}: For the task of entailment generation, we use the Stanford Natural Language Inference (SNLI) corpus~\cite{bowman2016snli} and MultiNLI~\cite{williams2017broad}. We use their entailment labeled pairs for our entailment generation task, following previous work~\cite{pasunuru2017multitask}. The combined SNLI and MultiNLI dataset has $302,879$ entailment pairs, out of which we use $276,720$ pairs for training, and the rest are divided into validation and test sets. 
\textbf{ParaNMT}: For the task of paraphrase generation, we use the back-translated paraphrase dataset provided by~\newcite{Wieting2017PushingTL}. The filtered version of the dataset has $5.3$ million pairs of paraphrases.\footnote{We chose ParaNMT over other paraphrase datasets (e.g. the phrase-to-phrase PPDB dataset~\cite{ganitkevitch2013ppdb}), because ParaNMT is a sentence-to-sentence dataset and hence is a more natural fit for sentence-level multi-task RNN-layer sharing with our sentence-to-sentence simplification task.} We use $99\%$ for training, and the rest are evenly divided into validation and test sets.

\paragraph{Evaluation Metrics}
\label{subsec:evaluation}

Following previous work~\cite{zhang2017dress}, we report all the standard evaluation metrics: SARI~\cite{Xu2016OptimizingSM}, FKGL~\cite{kincaid1975fkgl}, and BLEU~\cite{Papineni2002BleuAM}. However, several studies have shown that BLEU is poorly correlated w.r.t. simplicity~\cite{Zhu2010AMT,vstajner2015deeper,Xu2016OptimizingSM}. Moreover,~\newcite{shardlow2014survey} argues that FKGL~\cite{kincaid1975fkgl}, which measures readability of simpler output (lower is better), favors very short sentences even though longer/less coarse counterparts can be simpler. Further,~\newcite{Xu2016OptimizingSM} argues that BLEU tends to favor conservative systems that do not make many changes, and proposes SARI metric which explicitly measures the quality of words that are added and deleted.
SARI is shown to correlate well with human judgment in simplicity~\cite{Xu2016OptimizingSM},
and hence we primarily focus on this metric in our models' performance analysis.\footnote{We use the JOSHUA package for calculating SARI and BLEU score following~\newcite{zhang2017dress} and~\newcite{Xu2016OptimizingSM}. Our FKGL implementation is based on~\url{https://github.com/mmautner/readability.}} 
Further, we also do human evaluation based on: Fluency (`is the output grammatical and well formed?'), Adequacy (`to what extent is the meaning expressed in the original sentence preserved in the output?') and Simplicity (`is the output simpler than the original sentence?'), following guidelines suggested by~\newcite{Xu2016OptimizingSM} and~\newcite{zhang2017dress}.

\paragraph{Training Details}
\label{subsec:training}
All our soft/hard and layer-specific sharing decisions (\secref{sec:analysis}) were made on the validation/dev set. Our model selection (tuning) criteria is based on the average of our 3 metrics (SARI, BLEU, 1/FKGL) on the validation set. Please refer to the appendix for full training details (vocabulary overlap, mixing ratios and bandit sampler decay rates and reward, WikiLarge pre-training, etc.).

%% file: results.tex
\begin{table}[t]
\begin{minipage}[t]{.365\linewidth}
\begin{center}
\begin{small}
\begin{tabular}{|l|c|c|c|c|}
\hline
Models & BLEU & FKGL & SARI\\
\hline
\multicolumn{4}{|c|}{\textsc{Previous Work}}\\
\hline
PBMT-R & 18.19 & 7.59 & 15.77 \\
Hybrid & 14.46 & 4.01 & 30.00 \\
EncDecA & 21.70 & 5.11 & 24.12 \\
DRESS & 23.21 & 4.13 & 27.37 \\
DRESS-LS & 24.30 & 4.21 & 26.63 \\
\hline
\multicolumn{4}{|c|}{\textsc{Our Models}}\\
\hline
Baseline $\otimes$ & 23.72 & 3.25 & 28.31 \\
$\otimes$ + Ent. & 16.82 & 2.21 & 31.55 \\
$\otimes$ + Paraphr. & 16.29 & 2.03 & 31.71 \\
$\otimes$+Ent+Par & 11.86 & 1.38 & 32.98 \\
\hline
\end{tabular}
\end{small}
\end{center}
\vspace{-10pt}
\caption{Newsela (FKGL: lower is better). Note that SARI is the primary, human-correlated metric for sentence simplification~\cite{Xu2016OptimizingSM}.}
\vspace{-5pt}
\label{table:newsela_results}
\end{minipage}
\hfill
\begin{minipage}[t]{.595\linewidth}
\begin{center}
\begin{small}
\begin{tabular}{|l|c|c|c|c|c|c|c|}
\hline
& \multicolumn{3}{c|}{\textsc{WikiSmall}} & \multicolumn{3}{c|}{\textsc{WikiLarge}}\\
Models & BLEU & FKGL & SARI &  BLEU & FKGL & SARI\\
\hline
\multicolumn{7}{|c|}{\textsc{Previous Work}}\\
\hline
PBMT-R  & 46.31 & 11.42 & 15.97 & 81.11 & 8.33 & 38.56 \\
Hybrid  & 53.94 & 9.21 & 30.46 & 48.97 & 4.56 & 31.40 \\
SBMT-SARI  & - & - & -  & 73.08 & 7.29 & 39.96\\
EncDecA  & 47.93 & 11.35 & 13.61 & 88.85 & 8.41 & 35.66 \\
DRESS  & 34.53 & 7.48 & 27.48 & 77.18 & 6.58 & 37.08 \\
DRESS-LS  & 36.32 & 7.55 & 27.24 & 80.12 & 6.62 & 37.27 \\
\hline
\multicolumn{7}{|c|}{\textsc{Our Models}}\\
\hline
Baseline $\otimes$  & 36.18 & 7.69 & 25.67 & 82.37 & 7.84 & 36.68 \\
$\otimes$+Ent+Par  & 29.70 & 6.93 & 28.24 & 81.49 & 7.41 & 37.45 \\
\hline
\end{tabular}
\end{small}
\end{center}
\vspace{-8pt}
\caption{WikiSmall/Large results (FKGL: lower is better). Note that SARI is the primary, human-correlated metric for sentence simplification~\cite{Xu2016OptimizingSM}.
}
\label{table:wikilarge_results}
\end{minipage}
\vspace{-5pt}
\end{table}

\section{Results}
\label{label:results}
We evaluate our models on three datasets and via several automatic metrics plus human evaluation.\footnote{As described in Sec.~\ref{subsec:datasets}, Newsela is considered as a higher quality dataset for text simplification, and thus we report ablation-style results (e.g., 2-way multi-task models and different layer-sharing ablations) and human evaluation on Newsela (since Wikipedia datasets are automatically-aligned). Moreover, we report SARI, FKGL, and BLEU for completeness, but as described in Sec.~\ref{section:setup}, SARI is the primary human-correlated metric for sentence simplification.} 

\paragraph{Pointer Baseline}
First, we compare our pointer baseline with various previous works: PBMT-R~\cite{Wubben2012SentenceSB}, Hybrid~\cite{Narayan2014HybridSU}, SBMT-SARI~\cite{Xu2016OptimizingSM}\footnote{We borrow the SBMT-SARI results for WikiLarge from~\newcite{zhang2017dress}.}, and EncDecA, DRESS, and DRESS-LS~\cite{zhang2017dress}. As shown in Table~\ref{table:newsela_results}, our pointer baseline already achieves the best score in FKGL and the second-best score in SARI on Newsela, and also achieves overall comparable results on both WikiSmall and WikiLarge (see Table~\ref{table:wikilarge_results}).

\paragraph{Multi-Task Models}
We further improve our strong pointer-based sentence simplification baseline model by multi-task learning it with entailment and paraphrase generation. First, we show that our 2-way multi-task models with auxiliary tasks (entailment and paraphrase generation) are statistically significantly better than our pointer baseline and previous works in both SARI and FKGL on Newsela (see Table~\ref{table:newsela_results}).\footnote{Stat. significance is computed via bootstrap test~\cite{noreen1989computer,efron1994introduction}. Both our 2-way multi-task models are statistically significantly better in SARI and FKGL with $p<0.01$ w.r.t. our pointer baseline and previous works. Note the discussion in~\secref{subsec:evaluation} about why BLEU is not a good sentence simplification metric.} 
Next, Table~\ref{table:newsela_results} and Table~\ref{table:wikilarge_results} summarize the performance of our final 3-way multi-level, multi-task models with entailment generation and paraphrase generation on all three datasets.
Here, our 3-way multi-task models are statistically significantly better than our pointer baselines in both SARI and FKGL (with $p<0.01$) on Newsela and WikiSmall, and in SARI ($p<0.01$) on WikiLarge.
Also, our 3-way multi-task model is statistically significantly better than the 2-way multi-task models in SARI and FKGL with $p<0.01$ (see Table~\ref{table:newsela_results}). In Sec.~\ref{sec:analysis}, we further provide a set of detailed ablation experiments investigating the effects of different (higher-level versus lower-level) layer sharing methods and soft- vs. hard-sharing in our multi-level, multi-task models; and we show the superiority of our final choice of higher-level semantic sharing for entailment generation and lower-level lexico-syntactic sharing for paraphrase generation.

\begin{wraptable}[8]{r}{0.45\textwidth}
\begin{small}
\begin{center}
\vspace{-4pt}
\begin{tabular}{|l|c|c|c|}
\hline
Models & BLEU & FKGL & SARI \\
\hline
\multicolumn{4}{|c|}{\textsc{Newsela}}\\
\hline
Static Mixing Ratio  & 11.86 & 1.38 & 32.98 \\
Dynamic Mixing Ratio & 11.14 & 1.32  & 33.22 \\
\hline
\multicolumn{4}{|c|}{\textsc{WikiSmall}}\\
\hline
Static Mixing Ratio  & 29.70 &  6.93 & 28.24 \\
Dynamic Mixing Ratio & 27.23 &  5.86 & 29.58 \\
\hline
\end{tabular}
\end{center}
\vspace{-12pt}
\caption{Results on dynamic vs. static mixing ratio (FKGL: lower is better).}
\label{table:dynamic-mixing-results}
\end{small}
\end{wraptable}
\paragraph{Dynamic Mixing Ratio Models}
Finally, we present results on our 3-way multi-task model with the new approach of using `dynamic' mixing ratios based on multi-armed bandits sampling (see~\secref{subsec:active-sampling}). As shown in Table~\ref{table:dynamic-mixing-results}, this dynamic multi-task approach achieves a stat. significant improvement in SARI  as compared to the traditional fixed and manually-tuned mixing ratio based 3-way multi-task model: 33.22 vs. 32.98 ($p<0.05$) on Newsela, and 29.58 vs. 28.24 ($p<0.001$) on WikiSmall. Hence, this allows us to achieve not only equal, but in fact better results than the manual approach, while importantly avoiding the hassle of tuning on the large space of mixing ratios over several different tasks. In Sec.~\ref{sec:analysis}, we further provide ablation analysis to study whether the improvements come from the bandit learning this dynamic curriculum or from the bandit finding the final optimal mixing-ratio at the end of the sampling procedure (and also compare it to a random curriculum).

\begin{table}[t]
\begin{center}
\begin{small}
\begin{tabular}{|l|c|c|c|c|||c|c|c|}
\hline
 & \multicolumn{4}{c|||}{HUMAN EVALUATION} & \multicolumn{3}{c|}{MATCH-WITH-INPUT} \\
Models & Fluency & Adequacy & Simplicity & Average & BLEU (\%) & ROUGE (\%) & Exact Match (\%) \\
\hline
Ground-truth & 4.97 & 4.08 & 3.83 & 4.29 & 18.25 & 43.74 & 0.00 \\
\hdashline
Hybrid & 3.88 & 3.82 & 3.92 & 3.87 & 25.74 & 56.20 & 3.34 \\
DRESS-LS & 4.84 & 4.18 & 3.21 & 4.08 & 42.93 & 67.61 & 14.48 \\
Pointer Baseline & 4.61 & 3.94 & 3.99 & 4.18 & 30.80 & 60.56 & 10.68 \\
3-way Multi-task &  4.73 & 3.18 & 4.62 & 4.18 & 8.74 & 37.82 & 2.41 \\
\hline
\end{tabular}
\end{small}
\end{center}
\vspace{-10pt}
\caption{Human evaluation results (on left) and closeness-to-input source results (on right), for Newsela. In Sec.~\ref{label:results} `Human Evaluation', we discuss the issue of high adequacy scores for outputs that are very similar to the input (see right part of the table).
}
\vspace{-7pt}
\label{table:newsela-human-eval}
\end{table}

\paragraph{Human Evaluation}
We also perform an anonymized human study comparing our pointer baseline, our multi-task model, some previous works (Hybrid~\cite{Narayan2014HybridSU} and state-of-the-art DRESS-LS~\cite{zhang2017dress}), and ground-truth references (see left part of Table~\ref{table:newsela-human-eval}), based on fluency, adequacy, and simplicity (see Sec.~\ref{subsec:evaluation} for more details about these criteria) using 5-point Likert scale. We asked annotators to evaluate the models (randomly shuffled to anonymize model identity) based on $200$ samples from the representative and cleaner Newsela test set, and their scores are reported in Table~\ref{table:newsela-human-eval}.
Our 3-way multi-task model achieves a significantly higher ($p<0.001$) simplicity score compared to DRESS-LS, Hybrid, and our pointer baseline models. 
However, we next observe that our 3-way multi-task model has lower adequacy score as compared to DRESS-LS and the pointer model, but this is because our 3-way multi-task model focuses more strongly on simplification, which is the goal of the given task. 
Moreover, based on the overall average score of the three human evaluation criteria, our 3-way multi-task model is also significantly better ($p<0.03$) than the state-of-the-art DRESS-LS model (and $p<0.001$ w.r.t. Hybrid model).\footnote{Note that our multi-task model is stat. equal to our pointer baseline on the overall-average score, showing the available trade-off between systems that simplify conservatively vs. strongly, based on one's desired downstream task application. Also refer to the high `match-with-input' issue with the adequacy metric discussed next.} 
Also, on further investigation, we found that a problem with the adequacy metric is that it gets artificially high scores for output sentences which are exact match (or a very close match) with the input source sentence, i.e., they have very little simplification and hence almost fully retain the exact meaning. In the right part of Table~\ref{table:newsela-human-eval}, we analyzed the matching scores of the outputs from different models w.r.t. the source input text, based on BLEU, ROUGE~\cite{lin2004rouge} and exact match. First, this shows that the ground-truth sentence-simplification references are in fact (as expected) very different from the input source (0\% exact match, 18\% BLEU, 44\% ROUGE). Next, we find that our multi-task model also has low match-with-input scores (2\% exact match, 9\% BLEU, 38\% ROUGE), similar to the behavior of the ground-truth references. On the other hand, DRESS-LS (and pointer baseline) model is generating output sentences which are substantially closer to the input and hence is not making enough changes (14\% exact match, 43\% BLEU, 68\% ROUGE) as compared to the references (which explains their higher adequacy but lower simplicity scores).

%% file: analysis.tex
\section{Ablations and Analysis}
\label{sec:analysis}

In this section, we conduct several ablation analyses to study the different layer-sharing mechanisms (higher semantic vs. lower lexico-syntactic), soft- vs. hard-sharing, two dynamic multi-armed bandit approaches, and our model's learned entailment and paraphrasing skills. We also present and analyze some output examples from several models.\footnote{Since Newsela is considered as the more representative dataset for sentence simplification with lesser noise and human quality~\cite{xu2015problems,zhang2017dress}, we conduct our ablation studies on this dataset, but we observed similar patterns on the other two datasets as well.} Note that all our soft and layer sharing decisions were strictly made on the dev/validation set (see~\secref{subsec:training}).

\begin{wraptable}[7]{R}{0.51\textwidth}
    \vspace{-5pt}
\begin{small}
\begin{tabular}{|l|c|c|c|c|}
\hline
Models & BLEU & FKGL & SARI\\
\hline
\textbf{Final (High Ent + Low PP)} & \textbf{11.86} & \textbf{1.38} & 
\textbf{32.98} \\
\hline
Both lower-layer & 11.94 & 1.47 & 31.92 \\
Both higher-layer & 12.26 & 1.38 & 32.02 \\
Swapped (Low Ent + High PP) & 21.64 & 2.97 & 29.07 \\
\hline
Hard-sharing & 13.01 & 1.38 & 32.36 \\
\hline
\end{tabular}
\end{small}
\vspace{-12pt}
\caption{Multi-task layer ablation results on Newsela.}
\label{table:ablation_results}
\end{wraptable}

\paragraph{Different Layer Sharing Approaches}
We empirically show that our final multi-level layer sharing method (i.e., higher-level semantic layer sharing with entailment generation, while lower-level lexico-syntactic layer sharing with paraphrase generation) performs better than the following alternative layer sharing methods: (1) both auxiliary tasks with high-level layer sharing, (2) both with low-level layer sharing, and (3) reverse/swapped sharing (i.e., lower-level layer sharing for entailment, and higher-level layer sharing for paraphrasing). Results in Table~\ref{table:ablation_results} show that our approach of high-level sharing for entailment generation and low-level sharing for paraphrase generation is statistically significantly better than all other alternative approaches in SARI ($p<0.01$) (and statistically better or equal in FKGL).

\paragraph{Soft- vs. Hard-Sharing}
In this work, we use soft-sharing instead of hard-sharing approach (benefits discussed in Sec.~\ref{subsec:multi-level-sharing-mechanism}) in all of our models. Table~\ref{table:ablation_results} also presents empirical results comparing soft- vs. hard-sharing on our final 3-way multi-task model, and we observe that soft-sharing is statistically significantly better than hard-sharing in SARI with $p<0.01$.

\paragraph{Quantitative Improvements in Entailment}
We employ a state-of-the-art entailment classifier~\cite{chen2017enhanced} to calculate the entailment probabilities of output sentence being entailed by the ground-truth.\footnote{For this entailment analysis, we use ground-truth output as premise instead of input source, because: (1) entailment w.r.t. input source can give artificially high scores even when the output doesn't simplify enough and just copies the source (see the discussion in Sec.~\ref{label:results} and Table~\ref{table:newsela-human-eval}); (2) By transitivity, if output is entailed by ground-truth, which in turn is entailed by source, then output should also be entailed by source (plus, we want the output to be closer to ground-truth than to input source).}
Table~\ref{table:entailment_paraphrase_analysis_results} summaries the average entailment scores for the Hybrid, DRESS-LS, Pointer baseline, and 2-way multi-task model (with entailment generation auxiliary task), showing that the 2-way multi-task model improves in the aspect of logical entailment ($p<0.001$), demonstrating the inference skill acquired by the simplification model via the auxiliary knowledge from the entailment generation task.

\paragraph{Quantitative Improvements in Paraphrasing}
We use the paraphrase classifier from~\newcite{wieting17revisiting} to compute the paraphrase probability score between the generated output and the input source. The results in Table~\ref{table:entailment_paraphrase_analysis_results} show that our 2-way multi-task model (with paraphrasing generation auxiliary task) is closer to the ground-truth in terms of the amount of paraphrasing (w.r.t. input) required by the sentence-simplification task, while the pointer baseline and previous models have higher scores due to higher amount of copying from input source (see `Match-with-Input' discussion in Sec.~\ref{label:results}, Table~\ref{table:newsela-human-eval}).

\paragraph{Addition/Deletion Operations}
We also measured the performance of the various models in terms of the addition and deletion operations using SARI's sub-operation scores computed w.r.t. both the ground-truth and source~\cite{Xu2016OptimizingSM}. Table~\ref{table:individual_scores_results} shows that our multi-task model is equal or better in terms of both operations. 

\begin{table}[t]
\begin{minipage}[t]{.55\linewidth}
\begin{small}
\begin{center}
\begin{tabular}{|l|c|c|}
\hline
Models & Entailment & Paraphrasing \\
\hline
Ground-truth & N/A & 62.1 \\
\hdashline
Hybrid &  34.8 & 74.1 \\
DRESS-LS &  30.7 & 77.9 \\
Pointer Baseline &  36.9 &  76.6 \\
2-way Multi-Task & 41.4 & 63.9 \\
\hline
\end{tabular}
\end{center}
\vspace{-10pt}
\caption{Analysis: Entailment and paraphrase classification results (avg. probability scores as \%) on Newsela.}
\label{table:entailment_paraphrase_analysis_results}
\end{small}
\end{minipage}
\hfill
\begin{minipage}[t]{.41\linewidth}
\begin{small}
\begin{center}
\begin{tabular}{|l|c|c|}
\hline
Models & Deletions & Additions \\
\hline
Hybrid &  95.18 & 0.000 \\
DRESS-LS &  85.37 & 0.047 \\
Pointer Baseline &  88.91 &  0.026 \\
3-way Multi-Task & 97.54 & 0.049 \\
\hline
\end{tabular}
\end{center}
\vspace{-10pt}
\caption{Analysis: SARI's sub-operation scores on Newsela dataset.}
\label{table:individual_scores_results}
\end{small}
\end{minipage}
\vspace{-7pt}
\end{table}

\paragraph{Two Multi-Armed-Bandit Approaches}
As described in Sec.~\ref{subsec:active-sampling}, our multi-armed bandit approach with dynamic mixing ratio during multi-task training learns a sufficiently good curriculum to improve the sentence simplification task (see Sec.~\ref{label:results}). Here, we further show an ablation study on another alternative approach of using multi-armed bandits, where we record the last $10\%$ of the actions from the 
\begin{wrapfigure}[14]{r}{0.45\textwidth}
  \centering
  \vspace{0pt}
  \includegraphics[width=0.45\textwidth]{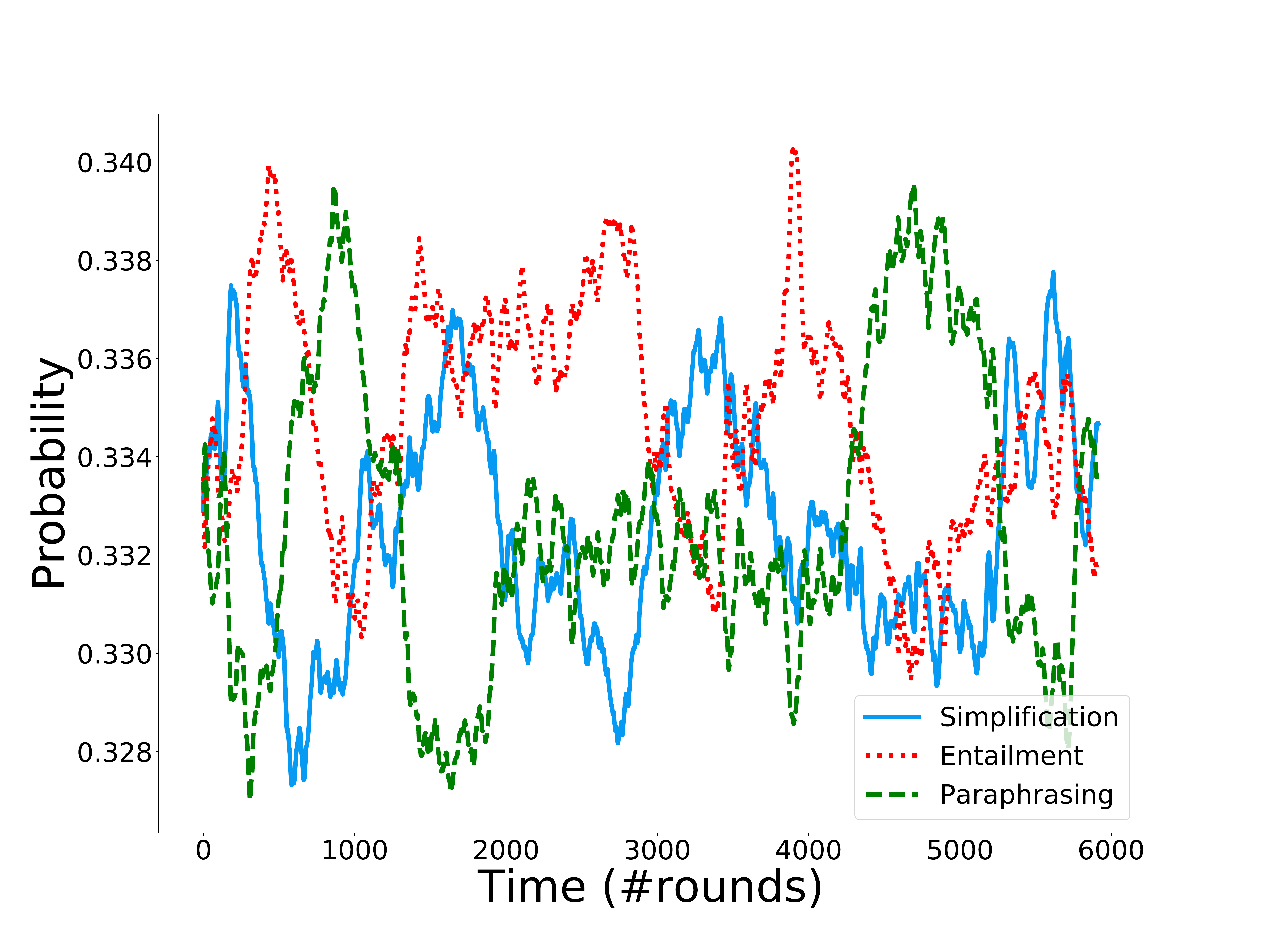}
  \vspace{-26pt}
  \caption{Task selection probability over training trajectory, predicted by bandit controller.}
	\label{fig:mab-visualization}
\end{wrapfigure}
bandit controller\footnote{We choose the last $10\%$ to avoid the noisy action-value estimates at the start of the training.}, then calculate the corresponding mixing ratio based on this 10\%, and run another independent model from scratch with this fixed mixing ratio. We found that the curriculum-style dynamic switching of tasks is in fact very effective as compared to this other 2-stage approach ($33.22$ versus $32.58$ in SARI with $p<0.01$). This is intuitive because the dynamic switching of tasks during multi-task training allows the model to choose the best next task to run based on the current state (as well as the previous curriculum path) of the model, as opposed to a fixed/static single mixing ratio for the full training period. 
In Fig.~\ref{fig:mab-visualization}, we visualize the (moving averages of) probabilities of selecting each task, which shows that in the 0-1000 \#rounds range, the bandit initially gives higher weight to the main task, but gradually redistributes the probabilities to the auxiliary tasks; and beyond 1000 \#rounds, it then alternates switching among the three different tasks periodically. 
We also experimented with replacing the bandit controller with random task choices, and our bandit-controller achieves statistically significantly better results than this approach in both SARI and FKGL with $p<0.01$, which shows that the path learned by the bandit controller is meaningful.

\begin{wrapfigure}[14]{R}{0.57\textwidth}
    \vspace{-16pt}
\begin{small}
\begin{tabularx}{\linewidth}{|X|l|}
\hline
\textbf{Input:} \emph{he put henson in charge of escorting his slaves to his brother 's kentucky plantation .} \\
\hdashline
\textbf{Reference:} \emph{he sent henson to take his slaves to kentucky .}\\
\textbf{DRESS-LS:} \emph{he put henson in charge of escorting his slaves to his brother 's kentucky plantation .} \\
\textbf{Baseline:} \emph{he put his slaves to his brother 's kentucky plantation .} \\
\textbf{Multi-Task:} \emph{he put henson in charge of escorting .}\\
\hline
\hline
\textbf{Input:} \emph{northern states did not allow slavery , but escaped slaves were returned to their owners as property , so henson would have to flee to canada to be free .} \\
\hdashline
\textbf{Reference:} \emph{states in the north did not allow slavery .}\\
\textbf{DRESS-LS:} \emph{southern states did not allow slavery , but the guatemalans were returned to their owners as property .} \\
\textbf{Baseline:} \emph{he slaves were returned to their owners as property .} \\
\textbf{Multi-Task:} \emph{northern states did not allow slavery .}\\
\hline
\end{tabularx}
 \vspace{-12pt}
\caption{Output examples comparing DRESS-LS, our pointer baseline, and multi-task model.}
\label{fig:examples}
\vspace{-5pt}
\end{small}
\end{wrapfigure}
\paragraph{Multi-Task Learning vs. Data Augmentation}
To verify that our improvements come indeed from the auxiliary tasks' specific character/capabilities and not just due to adding more data, we separately trained word embeddings on each auxiliary dataset (i.e., SNLI+MultiNLI and ParaNMT) and incorporated them into the primary simplification model. We found that both our 2-way multi-task models perform stat. significantly better than these models (which use the auxiliary word-embeddings), suggesting that merely adding more data is not enough. Moreover, Table~\ref{table:ablation_results} shows that only specific intuitive (syntactic vs. semantic) layer sharing between the primary and auxiliary tasks helps results and not just adding data.

\paragraph{Output Examples}
Fig.~\ref{fig:examples} shows two output examples comparing DRESS-LS, pointer baseline, and multi-task models (and reference). We see that our multi-task model simplifies the input appropriately (similar extent to reference) while also keeping reasonably important information from the source. The pointer baseline and the DRESS-LS models simplify to a lesser extent and keep much more of the original input (as also suggested by our match-with-input investigation in Table~\ref{table:newsela-human-eval}).

%% file: conclusions.tex
\section{Conclusion}
We presented a multi-level, multi-task learning approach to incorporate natural language inference and paraphrasing knowledge into sentence simplification models, via soft sharing at higher-level semantic and lower-level lexico-syntactic levels. We also introduced a multi-armed bandits approach for learning a dynamic mixing ratio of tasks. We demonstrated strong simplification improvements on three standard datasets via automatic and human evaluation, and also discussed several ablation and analysis studies.

%% file: appendix.tex
\section{Appendix}
\subsection{Training Details}
All LSTMs use hidden state size of $256$. We train word vectors with embedding size of $128$ with random initialization. 
We use gradient clipped norm of $2.0$. Our model selection (tuning) criteria is based on the average of our 3 metrics (SARI, BLEU, 1/FKGL) on the validation set. The mixing ratios are $\alpha_{ss}{:}\alpha_{eg}{:}\alpha_{pp} = 6{:}1{:}3$ for Newsela, $6{:}1{:}3$ for WikiSmall, and $7{:}2{:}1$ for WikiLarge. The soft-sharing coefficient $\lambda$ is set such that we balance the cross-entropy and regularization losses (at convergence),
which is $5 \times 10^{-6}$ for Newsela, $1 \times 10^{-6}$ WikiSmall, and $1 \times 10^{-5}$ for WikiLarge. We train models from scratch for Newsela and WikiSmall (using Adam~\cite{Kingma2014AdamAM} optimizer with learning rate of $0.002$ and $0.0015$, respectively). However, because of the large size and computation overhead for WikiLarge, we first pre-train both main and auxiliary models on their own domain until they reach $90\%$ convergence, and use these models to initialize the multi-task models, and set the learning rate to $1/10$ of its original default value ($0.001$). We set the decay rate $\alpha$ in the bandit controller to be $0.3$. We use the  negative validation loss as the reward at each sampling step to the bandit algorithm. The validation loss is divided by two as a smoothing technique.\footnote{This constant serves the same purpose as the temperature variable in the softmax function.} All our soft/hard and layer-specific sharing decisions (\secref{sec:analysis}) were made on the validation/dev set.
We follow previous work~\cite{zhang2017dress} in their pre-processing and post-processing of named entities.
We capped vocabulary size to be $50\textrm{K}$ and replaced less frequent words with $\textrm{UNK}$ token.\footnote{We measured the vocabulary overlap between the main and auxiliary tasks, and found that ``word-form-overlap'' (percentage of unique word types in auxiliary task that also appear in the main task) to be $40.7\%$ (entailment) and $41.0\%$ (paraphrase), and ``word-count-overlap'' (percentage of words in auxiliary task that also appear in the main task, based on token frequency counts) to be $95.2\%$ (entailment) and $94.9\%$ (paraphrase). Hence, this suggests that only rare words (which make up for very few counts) aren't considered in training process, and our pointer mechanism handles these extra UNK words by copying the actual word-form from the source to the output.} Unlike previous work~\cite{zhang2017dress}, we do not use $\textrm{UNK}$-replacement at test time, but instead rely on our pointer-copy mechanism. We use beam search with beam size of $5$. All other details provided in our released code.

%% file: main.bbl
\begin{thebibliography}{}

\bibitem[\protect\citename{Amancio and Specia}2014]{amancio2014analysis}
Marcelo Amancio and Lucia Specia.
\newblock 2014.
\newblock An analysis of crowdsourced text simplifications.
\newblock In {\em Proceedings of the 3rd Workshop on Predicting and Improving
  Text Readability for Target Reader Populations (PITR)}, pages 123--130.

\bibitem[\protect\citename{Auer \bgroup et al.\egroup }2002a]{auer2002finite}
Peter Auer, Nicolo Cesa-Bianchi, and Paul Fischer.
\newblock 2002a.
\newblock Finite-time analysis of the multiarmed bandit problem.
\newblock {\em Machine learning}, 47(2-3):235--256.

\bibitem[\protect\citename{Auer \bgroup et al.\egroup
  }2002b]{auer2002nonstochastic}
Peter Auer, Nicolo Cesa-Bianchi, Yoav Freund, and Robert~E Schapire.
\newblock 2002b.
\newblock The nonstochastic multiarmed bandit problem.
\newblock {\em SIAM journal on computing}, 32(1):48--77.

\bibitem[\protect\citename{Bahdanau \bgroup et al.\egroup
  }2015]{bahdanau2014attention}
Dzmitry Bahdanau, Kyunghyun Cho, and Yoshua Bengio.
\newblock 2015.
\newblock Neural machine translation by jointly learning to align and
  translate.
\newblock In {\em ICLR}.

\bibitem[\protect\citename{Barzilay and McKeown}2001]{barzilay2001extracting}
Regina Barzilay and Kathleen~R McKeown.
\newblock 2001.
\newblock Extracting paraphrases from a parallel corpus.
\newblock In {\em Proceedings of the 39th annual meeting on Association for
  Computational Linguistics}, pages 50--57. Association for Computational
  Linguistics.

\bibitem[\protect\citename{Belinkov \bgroup et al.\egroup
  }2017]{belinkov2017neural}
Yonatan Belinkov, Nadir Durrani, Fahim Dalvi, Hassan Sajjad, and James Glass.
\newblock 2017.
\newblock What do neural machine translation models learn about morphology?
\newblock {\em arXiv preprint arXiv:1704.03471}.

\bibitem[\protect\citename{Berant and Liang}2014]{berant2014semantic}
Jonathan Berant and Percy Liang.
\newblock 2014.
\newblock Semantic parsing via paraphrasing.
\newblock In {\em ACL (1)}, pages 1415--1425.

\bibitem[\protect\citename{Bowman \bgroup et al.\egroup }2015]{bowman2016snli}
Samuel~R Bowman, Gabor Angeli, Christopher Potts, and Christopher~D Manning.
\newblock 2015.
\newblock A large annotated corpus for learning natural language inference.
\newblock In {\em EMNLP}.

\bibitem[\protect\citename{Bubeck \bgroup et al.\egroup
  }2012]{bubeck2012regret}
S{\'e}bastien Bubeck, Nicolo Cesa-Bianchi, et~al.
\newblock 2012.
\newblock Regret analysis of stochastic and nonstochastic multi-armed bandit
  problems.
\newblock {\em Foundations and Trends{\textregistered} in Machine Learning},
  5(1):1--122.

\bibitem[\protect\citename{Carroll \bgroup et al.\egroup
  }1999]{carroll1999simplifying}
John~A Carroll, Guido Minnen, Darren Pearce, Yvonne Canning, Siobhan Devlin,
  and John Tait.
\newblock 1999.
\newblock Simplifying text for language-impaired readers.
\newblock In {\em EACL}, pages 269--270.

\bibitem[\protect\citename{Caruana}1998]{caruana1998multitask}
Rich Caruana.
\newblock 1998.
\newblock Multitask learning.
\newblock In {\em Learning to learn}, pages 95--133. Springer.

\bibitem[\protect\citename{Chandrasekar \bgroup et al.\egroup
  }1996]{chandrasekar1996motivations}
Raman Chandrasekar, Christine Doran, and Bangalore Srinivas.
\newblock 1996.
\newblock Motivations and methods for text simplification.
\newblock In {\em Proceedings of the 16th conference on Computational
  linguistics-Volume 2}, pages 1041--1044. Association for Computational
  Linguistics.

\bibitem[\protect\citename{Chapelle and Li}2011]{chapelle2011empirical}
Olivier Chapelle and Lihong Li.
\newblock 2011.
\newblock An empirical evaluation of thompson sampling.
\newblock In {\em Advances in neural information processing systems}, pages
  2249--2257.

\bibitem[\protect\citename{Chen \bgroup et al.\egroup }2017]{chen2017enhanced}
Qian Chen, Xiaodan Zhu, Zhen-Hua Ling, Si~Wei, Hui Jiang, and Diana Inkpen.
\newblock 2017.
\newblock Enhanced lstm for natural language inference.
\newblock In {\em Proceedings of the 55th Annual Meeting of the Association for
  Computational Linguistics (Volume 1: Long Papers)}, volume~1, pages
  1657--1668.

\bibitem[\protect\citename{Clarke and Lapata}2006]{clarke2006models}
James Clarke and Mirella Lapata.
\newblock 2006.
\newblock Models for sentence compression: A comparison across domains,
  training requirements and evaluation measures.
\newblock In {\em Proceedings of the 21st International Conference on
  Computational Linguistics and the 44th annual meeting of the Association for
  Computational Linguistics}, pages 377--384. Association for Computational
  Linguistics.

\bibitem[\protect\citename{Collobert and Weston}2008]{collobert2008unified}
Ronan Collobert and Jason Weston.
\newblock 2008.
\newblock A unified architecture for natural language processing: Deep neural
  networks with multitask learning.
\newblock In {\em Proceedings of the 25th international conference on Machine
  learning}, pages 160--167. ACM.

\bibitem[\protect\citename{Coster and Kauchak}2011]{coster2011learning}
William Coster and David Kauchak.
\newblock 2011.
\newblock Learning to simplify sentences using wikipedia.
\newblock In {\em Proceedings of the workshop on monolingual text-to-text
  generation}, pages 1--9. Association for Computational Linguistics.

\bibitem[\protect\citename{Dagan \bgroup et al.\egroup }2006]{dagan2006pascal}
Ido Dagan, Oren Glickman, and Bernardo Magnini.
\newblock 2006.
\newblock The pascal recognising textual entailment challenge.
\newblock In {\em Machine learning challenges. evaluating predictive
  uncertainty, visual object classification, and recognising tectual
  entailment}, pages 177--190. Springer.

\bibitem[\protect\citename{Devlin}1999]{devlin1999simplifying}
Siobhan~Lucy Devlin.
\newblock 1999.
\newblock {\em Simplifying natural language for aphasic readers.}
\newblock {Ph.D.} thesis, University of Sunderland.

\bibitem[\protect\citename{Efron and Tibshirani}1994]{efron1994introduction}
Bradley Efron and Robert~J Tibshirani.
\newblock 1994.
\newblock {\em An introduction to the bootstrap}.
\newblock CRC press.

\bibitem[\protect\citename{Evans \bgroup et al.\egroup
  }2014]{evans2014evaluation}
Richard Evans, Constantin Orasan, and Iustin Dornescu.
\newblock 2014.
\newblock An evaluation of syntactic simplification rules for people with
  autism.
\newblock In {\em Proceedings of the 3rd Workshop on Predicting and Improving
  Text Readability for Target Reader Populations (PITR)}, pages 131--140.

\bibitem[\protect\citename{Fader \bgroup et al.\egroup
  }2013]{fader2013paraphrase}
Anthony Fader, Luke Zettlemoyer, and Oren Etzioni.
\newblock 2013.
\newblock Paraphrase-driven learning for open question answering.
\newblock In {\em Proceedings of the 51st Annual Meeting of the Association for
  Computational Linguistics (Volume 1: Long Papers)}, volume~1, pages
  1608--1618.

\bibitem[\protect\citename{Filippova and Strube}2008]{filippova2008dependency}
Katja Filippova and Michael Strube.
\newblock 2008.
\newblock Dependency tree based sentence compression.
\newblock In {\em Proceedings of the Fifth International Natural Language
  Generation Conference}, pages 25--32. Association for Computational
  Linguistics.

\bibitem[\protect\citename{Ganitkevitch \bgroup et al.\egroup
  }2013]{ganitkevitch2013ppdb}
Juri Ganitkevitch, Benjamin Van~Durme, and Chris Callison-Burch.
\newblock 2013.
\newblock Ppdb: The paraphrase database.
\newblock In {\em HLT-NAACL}, pages 758--764.

\bibitem[\protect\citename{Girshick}2015]{girshick2015fast}
Ross Girshick.
\newblock 2015.
\newblock Fast r-cnn.
\newblock In {\em Proceedings of the IEEE international conference on computer
  vision}, pages 1440--1448.

\bibitem[\protect\citename{Glava{\v{s}} and
  {\v{S}}tajner}2015]{glavavs2015simplifying}
Goran Glava{\v{s}} and Sanja {\v{S}}tajner.
\newblock 2015.
\newblock Simplifying lexical simplification: Do we need simplified corpora.
\newblock In {\em Proceedings of the 53rd Annual Meeting of the Association for
  Computational Linguistics and the 7th International Joint Conference on
  Natural Language Processing}, volume~2, pages 63--68.

\bibitem[\protect\citename{Graves \bgroup et al.\egroup
  }2017]{graves2017automated}
Alex Graves, Marc~G Bellemare, Jacob Menick, Remi Munos, and Koray Kavukcuoglu.
\newblock 2017.
\newblock Automated curriculum learning for neural networks.
\newblock {\em arXiv preprint arXiv:1704.03003}.

\bibitem[\protect\citename{Guo \bgroup et al.\egroup }2018]{han2017multitask}
Han Guo, Ramakanth Pasunuru, and Mohit Bansal.
\newblock 2018.
\newblock Soft, layer-specific multi-task summarization with entailment and
  question generation.
\newblock In {\em Proceedings of ACL}.

\bibitem[\protect\citename{Harabagiu and Hickl}2006]{harabagiu2006methods}
Sanda Harabagiu and Andrew Hickl.
\newblock 2006.
\newblock Methods for using textual entailment in open-domain question
  answering.
\newblock In {\em ACL}, pages 905--912.

\bibitem[\protect\citename{Hashimoto \bgroup et al.\egroup
  }2017]{hashimoto2017ajm}
Kazuma Hashimoto, Caiming Xiong, Yoshimasa Tsuruoka, and Richard Socher.
\newblock 2017.
\newblock A joint many-task model: Growing a neural network for multiple nlp
  tasks.
\newblock In {\em EMNLP}.

\bibitem[\protect\citename{Horn \bgroup et al.\egroup }2014]{horn2014learning}
Colby Horn, Cathryn Manduca, and David Kauchak.
\newblock 2014.
\newblock Learning a lexical simplifier using wikipedia.
\newblock In {\em ACL (2)}, pages 458--463.

\bibitem[\protect\citename{Houthooft \bgroup et al.\egroup
  }2016]{houthooft2016vime}
Rein Houthooft, Xi~Chen, Yan Duan, John Schulman, Filip De~Turck, and Pieter
  Abbeel.
\newblock 2016.
\newblock Vime: Variational information maximizing exploration.
\newblock In {\em Advances in Neural Information Processing Systems}, pages
  1109--1117.

\bibitem[\protect\citename{Hwang \bgroup et al.\egroup
  }2015]{hwang2015aligning}
William Hwang, Hannaneh Hajishirzi, Mari Ostendorf, and Wei Wu.
\newblock 2015.
\newblock Aligning sentences from standard wikipedia to simple wikipedia.
\newblock In {\em NAACL-HLT}, pages 211--217.

\bibitem[\protect\citename{Inui \bgroup et al.\egroup }2003]{inui2003text}
Kentaro Inui, Atsushi Fujita, Tetsuro Takahashi, Ryu Iida, and Tomoya Iwakura.
\newblock 2003.
\newblock Text simplification for reading assistance: a project note.
\newblock In {\em Proceedings of the second international workshop on
  Paraphrasing-Volume 16}, pages 9--16. Association for Computational
  Linguistics.

\bibitem[\protect\citename{Jimenez \bgroup et al.\egroup
  }2014]{jimenez2014unal}
Sergio Jimenez, George Duenas, Julia Baquero, Alexander Gelbukh, Av~Juan~Dios
  B{\'a}tiz, and Av~Mendiz{\'a}bal.
\newblock 2014.
\newblock {UNAL-NLP}: Combining soft cardinality features for semantic textual
  similarity, relatedness and entailment.
\newblock In {\em In {SemEval}}, pages 732--742.

\bibitem[\protect\citename{Johnson \bgroup et al.\egroup
  }2016]{johnson2016google}
Melvin Johnson, Mike Schuster, Quoc~V Le, Maxim Krikun, Yonghui Wu, Zhifeng
  Chen, Nikhil Thorat, Fernanda Vi{\'e}gas, Martin Wattenberg, Greg Corrado,
  et~al.
\newblock 2016.
\newblock Google's multilingual neural machine translation system: enabling
  zero-shot translation.
\newblock {\em arXiv preprint arXiv:1611.04558}.

\bibitem[\protect\citename{Kaelbling \bgroup et al.\egroup
  }1996]{kaelbling1996reinforcement}
Leslie~Pack Kaelbling, Michael~L Littman, and Andrew~W Moore.
\newblock 1996.
\newblock Reinforcement learning: A survey.
\newblock {\em Journal of artificial intelligence research}, 4:237--285.

\bibitem[\protect\citename{Kaiser \bgroup et al.\egroup }2017]{kaiser2017one}
Lukasz Kaiser, Aidan~N Gomez, Noam Shazeer, Ashish Vaswani, Niki Parmar, Llion
  Jones, and Jakob Uszkoreit.
\newblock 2017.
\newblock One model to learn them all.
\newblock {\em arXiv preprint arXiv:1706.05137}.

\bibitem[\protect\citename{Kaji \bgroup et al.\egroup }2002]{kaji2002verb}
Nobuhiro Kaji, Daisuke Kawahara, Sadao Kurohash, and Satoshi Sato.
\newblock 2002.
\newblock Verb paraphrase based on case frame alignment.
\newblock In {\em Proceedings of the 40th Annual Meeting on Association for
  Computational Linguistics}, pages 215--222. Association for Computational
  Linguistics.

\bibitem[\protect\citename{Kauchak}2013]{Kauchak2013ImprovingTS}
David Kauchak.
\newblock 2013.
\newblock Improving text simplification language modeling using unsimplified
  text data.
\newblock In {\em ACL (1)}, pages 1537--1546.

\bibitem[\protect\citename{Kincaid \bgroup et al.\egroup
  }1975]{kincaid1975fkgl}
J~Peter Kincaid, Robert~P Fishburne~Jr, Richard~L Rogers, and Brad~S Chissom.
\newblock 1975.
\newblock Derivation of new readability formulas (automated readability index,
  fog count and flesch reading ease formula) for navy enlisted personnel.
\newblock Technical report, Naval Technical Training Command Millington TN
  Research Branch.

\bibitem[\protect\citename{Kingma and Ba}2014]{Kingma2014AdamAM}
Diederik~P. Kingma and Jimmy Ba.
\newblock 2014.
\newblock Adam: A method for stochastic optimization.
\newblock {\em CoRR}, abs/1412.6980.

\bibitem[\protect\citename{Klebanov \bgroup et al.\egroup
  }2004]{klebanov2004text}
Beata~Beigman Klebanov, Kevin Knight, and Daniel Marcu.
\newblock 2004.
\newblock Text simplification for information-seeking applications.
\newblock {\em Lecture Notes in Computer Science}, pages 735--747.

\bibitem[\protect\citename{Knight and Marcu}2002]{knight2002summarization}
Kevin Knight and Daniel Marcu.
\newblock 2002.
\newblock Summarization beyond sentence extraction: A probabilistic approach to
  sentence compression.
\newblock {\em Artificial Intelligence}, 139(1):91--107.

\bibitem[\protect\citename{Koehn \bgroup et al.\egroup }2007]{koehn2007moses}
Philipp Koehn, Hieu Hoang, Alexandra Birch, Chris Callison-Burch, Marcello
  Federico, Nicola Bertoldi, Brooke Cowan, Wade Shen, Christine Moran, Richard
  Zens, et~al.
\newblock 2007.
\newblock Moses: Open source toolkit for statistical machine translation.
\newblock In {\em Proceedings of the 45th annual meeting of the ACL on
  interactive poster and demonstration sessions}, pages 177--180. Association
  for Computational Linguistics.

\bibitem[\protect\citename{Lai and Hockenmaier}2014]{lai2014illinois}
Alice Lai and Julia Hockenmaier.
\newblock 2014.
\newblock Illinois-lh: A denotational and distributional approach to semantics.
\newblock {\em Proc. SemEval}, 2:5.

\bibitem[\protect\citename{Lin}2004]{lin2004rouge}
Chin-Yew Lin.
\newblock 2004.
\newblock Rouge: A package for automatic evaluation of summaries.
\newblock {\em Text Summarization Branches Out}.

\bibitem[\protect\citename{Luong \bgroup et al.\egroup }2015]{luong2015multi}
Minh-Thang Luong, Quoc~V Le, Ilya Sutskever, Oriol Vinyals, and Lukasz Kaiser.
\newblock 2015.
\newblock Multi-task sequence to sequence learning.
\newblock {\em arXiv preprint arXiv:1511.06114}.

\bibitem[\protect\citename{Narayan and Gardent}2014]{Narayan2014HybridSU}
Shashi Narayan and Claire Gardent.
\newblock 2014.
\newblock Hybrid simplification using deep semantics and machine translation.
\newblock In {\em ACL}.

\bibitem[\protect\citename{Noreen}1989]{noreen1989computer}
Eric~W Noreen.
\newblock 1989.
\newblock {\em Computer-intensive methods for testing hypotheses}.
\newblock Wiley New York.

\bibitem[\protect\citename{Papineni \bgroup et al.\egroup
  }2002]{Papineni2002BleuAM}
Kishore Papineni, Salim Roukos, Todd Ward, and Wei-Jing Zhu.
\newblock 2002.
\newblock Bleu: a method for automatic evaluation of machine translation.
\newblock In {\em ACL}.

\bibitem[\protect\citename{Parikh \bgroup et al.\egroup
  }2016]{parikh2016decomposable}
Ankur~P Parikh, Oscar T{\"a}ckstr{\"o}m, Dipanjan Das, and Jakob Uszkoreit.
\newblock 2016.
\newblock A decomposable attention model for natural language inference.
\newblock {\em arXiv preprint arXiv:1606.01933}.

\bibitem[\protect\citename{Pasunuru and Bansal}2017]{pasunuru2017multitask}
Ramakanth Pasunuru and Mohit Bansal.
\newblock 2017.
\newblock Multi-task video captioning with video and entailment generation.
\newblock In {\em Proceedings of ACL}.

\bibitem[\protect\citename{Pasunuru \bgroup et al.\egroup
  }2017]{Pasunuru2017TowardsIA}
Ramakanth Pasunuru, Han Guo, and Mohit Bansal.
\newblock 2017.
\newblock Towards improving abstractive summarization via entailment
  generation.
\newblock In {\em NFiS@EMNLP}.

\bibitem[\protect\citename{Petersen and Ostendorf}2007]{petersen2007text}
Sarah~E Petersen and Mari Ostendorf.
\newblock 2007.
\newblock Text simplification for language learners: a corpus analysis.
\newblock In {\em Workshop on Speech and Language Technology in Education}.

\bibitem[\protect\citename{Rello \bgroup et al.\egroup }2013]{rello2013impact}
Luz Rello, Ricardo Baeza-Yates, and Horacio Saggion.
\newblock 2013.
\newblock The impact of lexical simplification by verbal paraphrases for people
  with and without dyslexia.
\newblock In {\em International Conference on Intelligent Text Processing and
  Computational Linguistics}, pages 501--512. Springer.

\bibitem[\protect\citename{Ruder \bgroup et al.\egroup }2017]{ruder2017sluice}
Sebastian Ruder, Joachim Bingel, Isabelle Augenstein, and Anders S{\o}gaard.
\newblock 2017.
\newblock Sluice networks: Learning what to share between loosely related
  tasks.
\newblock {\em arXiv preprint arXiv:1705.08142}.

\bibitem[\protect\citename{See \bgroup et al.\egroup }2017]{see2017get}
Abigail See, Peter~J Liu, and Christopher~D Manning.
\newblock 2017.
\newblock Get to the point: Summarization with pointer-generator networks.
\newblock {\em arXiv preprint arXiv:1704.04368}.

\bibitem[\protect\citename{Shardlow}2014]{shardlow2014survey}
Matthew Shardlow.
\newblock 2014.
\newblock A survey of automated text simplification.
\newblock {\em International Journal of Advanced Computer Science and
  Applications}, 4(1):58--70.

\bibitem[\protect\citename{Sharma and Ravindran}2017]{sharma2017online}
Sahil Sharma and Balaraman Ravindran.
\newblock 2017.
\newblock Online multi-task learning using active sampling.
\newblock {\em CoRR}, abs/1702.06053.

\bibitem[\protect\citename{Siddharthan}2006]{SIDDHARTHAN2006SyntacticSA}
Advaith Siddharthan.
\newblock 2006.
\newblock Syntactic simplification and text cohesion.
\newblock {\em Research on Language and Computation}, 4(1):77--109.

\bibitem[\protect\citename{Siddharthan}2014]{Siddharthan2014ASO}
Advaith Siddharthan.
\newblock 2014.
\newblock A survey of research on text simplification.
\newblock {\em ITL-International Journal of Applied Linguistics},
  165(2):259--298.

\bibitem[\protect\citename{Specia}2010]{specia2010translating}
Lucia Specia.
\newblock 2010.
\newblock Translating from complex to simplified sentences.
\newblock {\em Computational Processing of the Portuguese Language}, pages
  30--39.

\bibitem[\protect\citename{{\v{S}}tajner \bgroup et al.\egroup
  }2015]{vstajner2015deeper}
Sanja {\v{S}}tajner, Hannah B{\'e}chara, and Horacio Saggion.
\newblock 2015.
\newblock A deeper exploration of the standard pb-smt approach to text
  simplification and its evaluation.
\newblock In {\em Proceedings of the 53rd Annual Meeting of the Association for
  Computational Linguistics (ACL)}.

\bibitem[\protect\citename{Sutton and Barto}1998]{sutton1998reinforcement}
Richard~S Sutton and Andrew~G Barto.
\newblock 1998.
\newblock {\em Reinforcement learning: An introduction}, volume~1.
\newblock MIT press Cambridge.

\bibitem[\protect\citename{Vickrey and Koller}2008]{vickrey2008sentence}
David Vickrey and Daphne Koller.
\newblock 2008.
\newblock Sentence simplification for semantic role labeling.
\newblock In {\em ACL}, pages 344--352.

\bibitem[\protect\citename{Wieting and Gimpel}2017a]{Wieting2017PushingTL}
John Wieting and Kevin Gimpel.
\newblock 2017a.
\newblock Pushing the limits of paraphrastic sentence embeddings with millions
  of machine translations.
\newblock {\em CoRR}, abs/1711.05732.

\bibitem[\protect\citename{Wieting and Gimpel}2017b]{wieting17revisiting}
John Wieting and Kevin Gimpel.
\newblock 2017b.
\newblock Revisiting recurrent networks for paraphrastic sentence embeddings.
\newblock In {\em Proceedings of the Annual Meeting of the Association for
  Computational Linguistics}.

\bibitem[\protect\citename{Williams \bgroup et al.\egroup
  }2017]{williams2017broad}
Adina Williams, Nikita Nangia, and Samuel~R Bowman.
\newblock 2017.
\newblock A broad-coverage challenge corpus for sentence understanding through
  inference.
\newblock {\em arXiv preprint arXiv:1704.05426}.

\bibitem[\protect\citename{Woodsend and Lapata}2011]{Woodsend2011LearningTS}
Kristian Woodsend and Mirella Lapata.
\newblock 2011.
\newblock Learning to simplify sentences with quasi-synchronous grammar and
  integer programming.
\newblock In {\em Proceedings of the conference on empirical methods in natural
  language processing}, pages 409--420. Association for Computational
  Linguistics.

\bibitem[\protect\citename{Woodsend and Lapata}2014]{woodsend2014text}
Kristian Woodsend and Mirella Lapata.
\newblock 2014.
\newblock Text rewriting improves semantic role labeling.
\newblock {\em Journal of Artificial Intelligence Research}, 51:133--164.

\bibitem[\protect\citename{Wubben \bgroup et al.\egroup
  }2012]{Wubben2012SentenceSB}
Sander Wubben, Antal van~den Bosch, and Emiel Krahmer.
\newblock 2012.
\newblock Sentence simplification by monolingual machine translation.
\newblock In {\em ACL}.

\bibitem[\protect\citename{Xu \bgroup et al.\egroup }2015]{xu2015problems}
Wei Xu, Chris Callison-Burch, and Courtney Napoles.
\newblock 2015.
\newblock Problems in current text simplification research: New data can help.
\newblock {\em Transactions of the Association of Computational Linguistics},
  3(1):283--297.

\bibitem[\protect\citename{Xu \bgroup et al.\egroup }2016]{Xu2016OptimizingSM}
Wei Xu, Courtney Napoles, Ellie Pavlick, Quanze Chen, and Chris Callison-Burch.
\newblock 2016.
\newblock Optimizing statistical machine translation for text simplification.
\newblock {\em Transactions of the Association for Computational Linguistics},
  4:401--415.

\bibitem[\protect\citename{Zeiler and Fergus}2014]{zeiler2014visualizing}
Matthew~D Zeiler and Rob Fergus.
\newblock 2014.
\newblock Visualizing and understanding convolutional networks.
\newblock In {\em European conference on computer vision}, pages 818--833.
  Springer.

\bibitem[\protect\citename{Zhang and Lapata}2017]{zhang2017dress}
Xingxing Zhang and Mirella Lapata.
\newblock 2017.
\newblock Sentence simplification with deep reinforcement learning.
\newblock {\em arXiv preprint arXiv:1703.10931}.

\bibitem[\protect\citename{Zhang \bgroup et al.\egroup
  }2015]{zhang2015exploiting}
Congle Zhang, Stephen Soderland, and Daniel~S Weld.
\newblock 2015.
\newblock Exploiting parallel news streams for unsupervised event extraction.
\newblock {\em Transactions of the Association for Computational Linguistics},
  3:117--129.

\bibitem[\protect\citename{Zhu \bgroup et al.\egroup }2010]{Zhu2010AMT}
Zhemin Zhu, Delphine Bernhard, and Iryna Gurevych.
\newblock 2010.
\newblock A monolingual tree-based translation model for sentence
  simplification.
\newblock In {\em Proceedings of the 23rd international conference on
  computational linguistics}, pages 1353--1361. Association for Computational
  Linguistics.

\end{thebibliography}
